\pdfoutput=1

\documentclass[11pt]{article}
\usepackage{authblk}

\usepackage{acl}

\usepackage{times}
\usepackage{latexsym}
\usepackage{graphicx}
\graphicspath{{./}}

\usepackage{tabularx}
\usepackage{setspace}
\usepackage{booktabs}
\usepackage{microtype}
\usepackage{inconsolata}
\usepackage{booktabs}
\usepackage{multirow}
\usepackage{enumitem}
\usepackage{makecell}
\usepackage{adjustbox}
\usepackage{tablefootnote}
\usepackage{color}
\usepackage{multirow}
\usepackage{adjustbox}
\usepackage{enumitem}
\usepackage{algorithmicx,algorithm}
\usepackage{amsmath}
\usepackage{geometry}
\usepackage{booktabs}
\usepackage{algpseudocode}
\usepackage{multirow}
\usepackage{amsmath}
\usepackage{booktabs}
\usepackage{float}
\usepackage{times}

\usepackage[T1]{fontenc}

\usepackage[utf8]{inputenc}

\usepackage{microtype}

\usepackage{inconsolata}

\title{Enhancing Dialogue State Tracking Models through LLM-backed User-Agents Simulation}

\author[1]{\textbf{Cheng Niu}}
\author[1]{\textbf{Xingguang Wang}}
\author[1]{\textbf{Xuxin Cheng}}
\author[1]{\textbf{Juntong Song}}
\author[2]{\textbf{Tong Zhang}}
\affil[1]{NewsBreak}
\affil[2]{University of Illinois Urbana-Champaign}
\affil[ ]{ {cheng.niu@newsbreak.com}}

\begin{document}
\maketitle
\begin{abstract}
Dialogue State Tracking (DST) is designed to monitor the evolving dialogue state in the conversations and plays a pivotal role in developing task-oriented dialogue systems. However, obtaining the annotated data for the DST task is usually a costly endeavor. In this paper, we focus on employing LLMs to generate dialogue data to reduce dialogue collection and annotation costs. Specifically, GPT-4 is used to simulate the user and agent interaction, generating thousands of dialogues annotated with DST labels. Then a two-stage fine-tuning on LLaMA 2 is performed on the generated data and the real data for the DST prediction. Experimental results on two public DST benchmarks show that with the generated dialogue data, our model performs better than the baseline trained solely on real data. In addition, our approach is also capable of adapting to the dynamic demands in real-world scenarios, generating dialogues in new domains swiftly. After replacing dialogue segments in any domain with the corresponding generated ones, the model achieves comparable performance to the model trained on real data.\footnote{All the source code, models, and generated dialogue data will be released after review for reproducibility.}
\end{abstract}
\section{Introduction}
Dialogue state tracking (DST) is a critical component of task-oriented dialogue systems, serving to track users' goals and system actions in the conversation and facilitate precise information handling for communicating with external APIs~\cite{henderson2014second,mrkvsic2017neural,zhang-etal-2023-monet,hudevcek2023large}. DST usually takes the form of key-value pairs, where the keys are denoted as slots which are defined in the system schema, outlining the specific information that the system aims to track or extract during the whole conversation~\cite{ren2018towards}. 

The design of DST models could be broadly categorized into two main types,  \textit{classification-based DST models} and \textit{generation-based DST models}. \textit{Classification-based models} select slot values from a set of candidates~\cite{ma-etal-2019-implicit,ye2021slot}, assuming that the dialogue ontology is pre-defined and hence lacking generalization capability~\cite{chen2020schema,wang2022luna}. \textit{Generation-based models} directly generate the slot values to handle unseen domains and values \cite{gao-etal-2019-dialog, gao-etal-2020-machine,lin-etal-2020-mintl, peng-etal-2021-soloist}. Recently, \citet{feng2023towards} proposes a new DST framework LDST based on LLaMA~\cite{touvron2023llama}. By using an instruction tuning method, LDST achieves performance on par with ChatGPT~\cite{chatgptPlayground}.

Despite DST showing promising results, a significant challenge is that the annotation of dialogues entails significant costs. Furthermore, the dynamic nature of real-world demands highlights the urgent need to quickly generate utterances for new domains. Compared to other types of NLP data, collecting authentic dialogue data is particularly challenging. This difficulty is partly due to the dialogues frequently containing personal or sensitive information, which complicates data collection and sharing efforts. In response to these challenges, and inspired by the recent advancements of large language models (LLMs)~\cite{llama2,autogpt2023,jablonka2023gpt,shen2023hugginggpt}, we explore the use of these models for generating annotated DST data for data augmentation. By leveraging LLM's cross-domain generation capability, we aim to create synthetic dialogues that can serve as replacements for manually annotated data, significantly reducing both financial cost and time constraints.

In this paper, we propose a \textbf{L}LM-backed \textbf{U}ser-\textbf{A}gents \textbf{S}imulation~(LUAS) algorithm to enhance DST. The process begins with the LLM generating a user profile that details the individual's preferences for various tasks. Following this initial step, the LLM is prompted to simulate a conversation between the user and the agent. In these simulations, the user simulator makes requests and seeks recommendations or assistance, while the agent responds by understanding the user's needs, providing suggestions, and taking appropriate actions. Through iterative conversations between the user and agent, complemented by a slot extractor also prompted by the LLM, we generate a substantial corpus of labeled, multi-turn dialogue data.


To verify the effectiveness of our approach and the quality of the generated data, experiments are conducted on two public DST datasets, MultiWOZ 2.2~\citep{woz2.2} and MultiWOZ 2.4~\citep{woz2.4}. Following ~\citet{llama2}, LLaMa 2 is finetuned with real data as a strong baseline. By using both the generated and the real data, finetuning LLaMa 2 can further improve the performance. Besides, by replacing dialogue segments of any domain with the generated data, the newly trained model achieves comparable performance to the model trained on the real data, which shows the capability of our method to meet the dynamic requirements of real-world scenarios, generating dialogues in new domains and preserving the promising performance.

In summary, the contributions of our work can be categorized into four aspects:
\begin{itemize}[leftmargin=*,itemsep=1pt,partopsep=0pt,topsep=0pt,parsep=0pt]
\item We propose a new framework that harnesses the power of GPT-4 to generate new labeled dialogue data, effectively reducing dialogue data collection and annotation costs. 
\item Experiment results on two datasets show the positive impact of the generated data on performance.
\item Our method can swiftly generate data in new domains while maintaining promising performance.
\item We believe that our approach holds promise for extension to other dialogue-related tasks.
\end{itemize}

\section{Related Work}
\subsection{Dialogue State Tracking}
Dialogue state tracking is an essential yet challenging task in task-oriented dialogue systems \citep{mrkvsic2017neural}. Recent DST models \citep{lee-etal-2021-dialogue, zhu-etal-2022-continual,yang2023multi,su2023choice,lesci-etal-2023-diable}, leveraging the different architectures and mechanisms, have convincingly demonstrated promising performance on several datasets \citep{budzianowski-etal-2018-multiwoz, woz2.1,woz2.2,woz2.3,woz2.4}. To ease the burden of dialogue collection and annotation, \citet{wu-etal-2019-transferable,zhou2023xqa} use few-shot learning to transfer to adapt existing models to the new domains.
Drawn by the recent achievement of LLMs, \citet{feng2023towards} leverages Low-Rank Adaptation~(LoRA)~\cite{hu2021lora} to fine-tune the foundation model, achieving the promising performance in DST. In this paper, we utilize GPT-4 to simulate user-agent conversations, and the obtained dialogue data significantly enhances DST.
\subsection{Data Augmentation by LLMs}
\begin{table*}[ht]
\centering
\resizebox{.98\linewidth}{!}{
\begin{tabular}{p{\textwidth}}
\toprule
DST: [\textbf{\textit{history}}], [\textbf{\textit{user\_utterance}}] $\rightarrow$ [\textbf{\textit{service}}], [\textbf{\textit{slot\_key}}], [\textbf{\textit{slot\_val}}]\\
\hline
You are a local guide online, primarily handling the local services like finding the user's place (such as attraction, hotel, train, restaurant, or hospital), calling taxis, contacting the police, or other convenient services. Your service is efficient and of high quality, earning widespread praise from the local community. Given the conversion history, your task is to help find what the user is looking for based on the whole conversion. Please output the current\_service based on the user's last utterance. And also please output all service information that needs to be paid attention to from the whole conversion. Here are the ``conversion history'': \{[\textbf{\textit{history}}]\} and the ``user's lastest utterance'': \{[\textbf{\textit{user\_utterance}}]\}. The output should be JSON-formatted like {{``current\_service'': \{[\textbf{\textit{service}}]\}, ``slots'': {\{``[\textbf{\textit{service}}]'': \{``[\textbf{\textit{slot\_key}}]'': \{[\textbf{\textit{slot\_val}}]\}\}\}}}}.\\ Please give your decision: \\
\bottomrule
\end{tabular}}
\caption{Proposed prompts to guide LLaMA 2 to generate JSON-formatted dialogue state predictions.}
\label{table:prompts}
\end{table*}

Data augmentation has shown remarkable effectiveness in various domains, including computer vision \cite{krizhevsky2012imagenet,shorten2019survey}, text classification \cite{zhang2015character,wei2019eda}, and speech recognition \cite{ko15_interspeech,park19e_interspeech}. 

In recent years, with the increasing prominence of LLMs, an increasing number of studies have begun to utilize LLMs for data augmentation.
\citet{kaddour2024synthetic} discovers that fine-tuning teacher LLMs to annotate unlabeled instances and generate new data points can notably boost the performance of downstream models.
\citet{yang2023refgpt} generates truthful and customized dialogues to reduce hallucation.
\citet{ulmer2024bootstrapping} compares the effectiveness of various filtering strategies for the generated dialogue quality and introduces new methods to benchmark finetuned dialogue system. But their work does not discuss the DST task. \citet{zekun2022} presented GPT-3 backed user-agent simulation system and showed positive results on DST task when the real data size is extremely small. Unlike \citet{zekun2022}, we abstract the common intentions of users and agents, crafting intent-specific prompts to ensure that the simulation adheres to task-oriented logic. This scheme enables the simulation to operate within a zero-shot setup, enhancing our approach's adaptability to new domains. Moreover, by implementing a two-stage fine-tuning process, our approach demonstrates superior performance compared to strong baselines, even when trained with the full size of real data.

\section{Method}
In this section, we will begin with the basic problem definition~($\S\ref{Problem Definition}$). Then, we introduce our proposed method, including fine-tuning LLaMA 2 to predict dialogue state~($\S\ref{Using LLaMA 2 to Predict Dialogue State}$) and utilizing GPT-4 for user-agent conversation simulation~($\S\ref{Employing GPT-4 to Simulate Comprehensive Dialogue Data}$). Finally, we present our two-stage fine-tuning strategy to use both generated and real data for DST~($\S\ref{Two-stage Fine-tuning Strategy}$).
\subsection{Problem Definition}
\label{Problem Definition}
A task-oriented dialogue involves a multi-turn conversation between a user $U$ and an agent $A$. Given a dialogue context $C_t = [U_1, A_1, ..., U_t, A_t]$ as the sequence of utterances up to turn $t$, the goal of DST is to predict the dialogue state $y_t$, which is defined as a collection of \textit{(slot, value)} pairs:
\begin{equation*}
    y_t = \{(s^i_t, v^i_t) \;|\; C_t \; , \forall s^i \in \mathcal{S}\}
\end{equation*}
where $\mathcal{S}$ denotes the set of the possible slots predefined in an ontology or schema. Following previous work~\cite{wang-etal-2023-divide}, the final slot is represented as the concatenation of the corresponding task domain and original slot, e.g., ``<hotel-area>''. The slots associated with each domain could be either categorical with a set of candidate values (e.g. <hotel-parking> = ``True'' / ``False''), or non-categorical, where the value is a span in the dialogue context (e.g. <hotel-name> = ``Alexander''). Note that if no information is provided in the dialogue regarding a specific slot, the associated value for that slot is set to ``NONE''.

\subsection{Using LLaMA 2 to Predict Dialogue State}
\label{Using LLaMA 2 to Predict Dialogue State}
\begin{figure*}[t]
\centering
\resizebox{.98\linewidth}{!}{
\includegraphics[width=\linewidth]{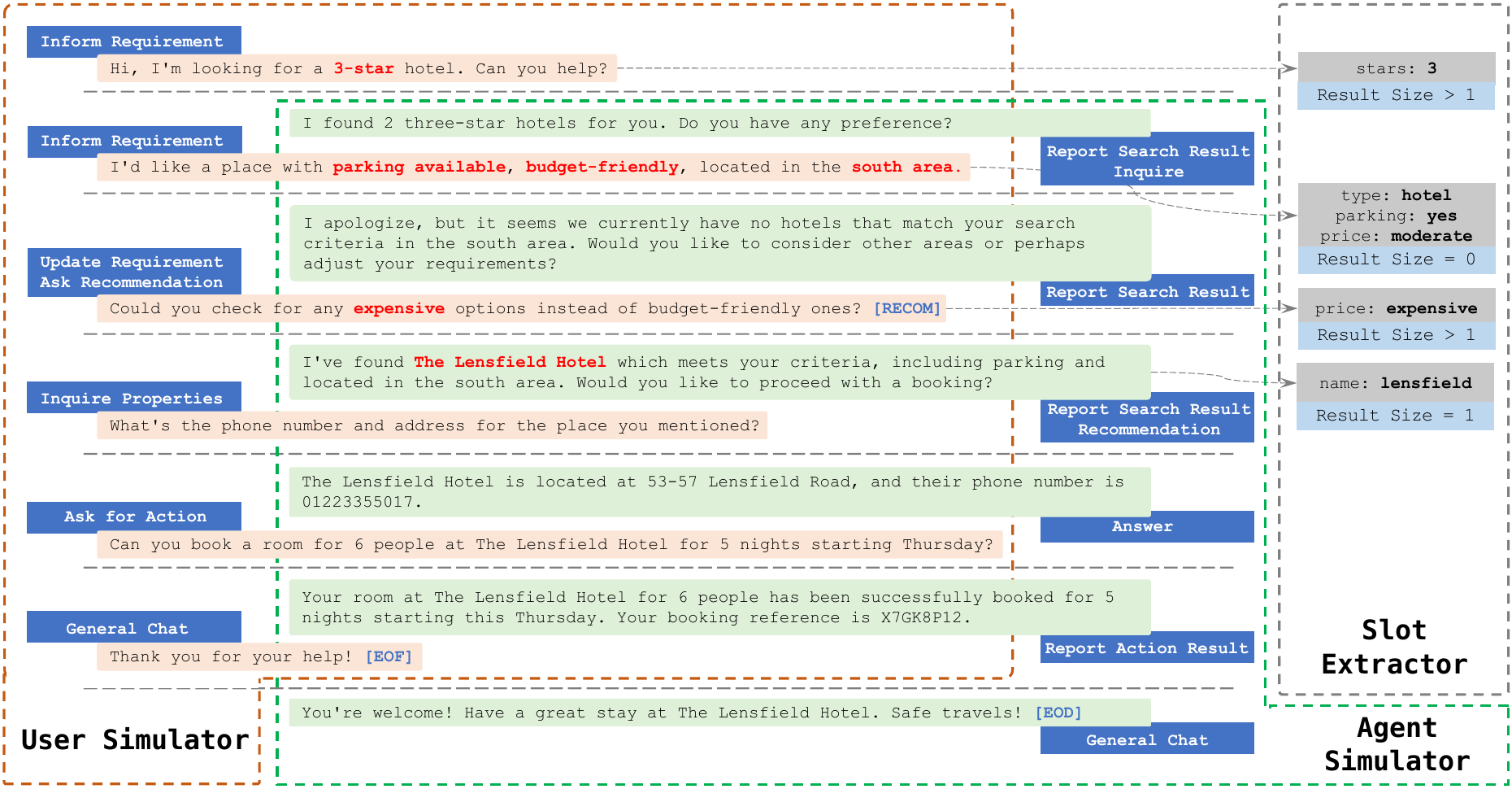}
}
\caption{The simulation process of our approach. The blue boxes are intentions for the user and the agent, the `[RECOM]', `[EOF]', and `[EOD]' are control identifiers.}
\label{fig:method}
\end{figure*}
We employ full-parameter fine-tuning on LLaMA 2 to predict dialogue states and employ pre-designed prompts to guide the LLaMA 2 model in generating predictions formatted in JSON. As demonstrated in Table \ref{table:prompts}, dialogue history and the user's latest utterance are fed into LLaMA 2, which then conducts the prediction of the entire conversation's intents and slot values. Specifically, predicted intents must fall within a predefined set, and predicted slots must align with the designated slots for the respective intents. We implement a schema to prevent the generation model from producing incoherent outputs and to enhance the overall quality and reliability of the outputs of LLaMA 2. The optimization is conducted through the utilization of cross-entropy. 

\subsection{User-Agent Dialogue Simulation backed by GPT-4}
\label{Employing GPT-4 to Simulate Comprehensive Dialogue Data}
As illustrated in Figure \ref{fig:method}, the dialogue simulation framework based on GPT-4 involves a multi-stage approach for producing labeled multi-turn dialogue data. In this arrangement, GPT-4 prompts two simulators, including the \textbf{user simulator} and the \textbf{agent simulator}, to engage in conversations aimed at completing specific dialogue tasks. Concurrently, GPT-4 also prompts a \textbf{slot extractor} to identify and extract all relevant slots throughout the entire conversation simulation process.

The details of the simulation generation process are outlined below, with all the prompts included in the appendix for reference.


\subsubsection{Simulation Process Overview}
\label{Overview}
Before initiating the dialogue, GPT-4 is prompted to create a user profile that outlines the individual's preferences across various tasks such as travel, accommodations, dining, and more. Each preference includes specific details like budget, travel distance, and other criteria. Following this setup, the user simulator begins interacting with the agent, presenting its requests and seeking recommendations or assistance with bookings and purchases. The agent, in turn, is prompted to delve into the user's needs, conduct searches for pertinent information, offer suggestions, and execute necessary actions. After each interaction, the user simulator evaluates how well their needs have been met, deciding whether to continue the conversation.

\subsubsection{User/Agent Intentions}
To effectively navigate the simulators through interactive tasks, we encounter the challenge of encoding complex dialogue logic within a single prompt. This task is demanding for both the user and the agent simulator. To simplify, we abstract the common intentions of users and agents, and craft prompts specifically for each unique intention of the user or agent. The detailed prompts of different intentions are shown in Appendix \ref{Prompts for Simulation}.

The user intentions are listed below:


\begin{itemize}[leftmargin=*,itemsep=1pt,partopsep=0pt,topsep=0pt,parsep=0pt]
\item \textbf{Inform Requirement}, the user informs their requirement to the agent.
\item \textbf{Update Requirement}, the user may update their requirements if the search result does not meet their criteria.
\item \textbf{Ask for Recommendation}, the user asks for a recommendation given a few candidates meeting their criteria. 
\item \textbf{Inquire Properties}, the user asks for some properties (e.g. address, etc.) of the candidates. 
\item \textbf{Ask for Action}, the user requires action after receiving the recommendation (e.g. making a reservation, etc.).
\item \textbf{General Chat}, other scenarios in the simulation, e.g. greeting or showing gratitude.
\end{itemize}

The agent intentions are listed below:
\begin{itemize}[leftmargin=*,itemsep=1pt,partopsep=0pt,topsep=0pt,parsep=0pt]
\item \textbf{Inquire}, ask the user's need and preference or seek the user's approval or confirmation.
\item \textbf{Report Search Results}, based on the user's preference, search the database and then make inquiries, recommendations, or reservations.
\item \textbf{Recommendation}, when more than one candidate meets users' search criteria, select the top candidate to recommend.
\item \textbf{Answer}, answer the user's inquiry about a recommendation from the agent.
\item \textbf{Report Action Result}, take action per the user's request and report the outcome of the action.
\item \textbf{General Chat}, other scenarios in the simulation, e.g. greetings or asking if there are any additional requirements to be addressed.
\end{itemize}

Besides natural language outputs, the simulators are also prompted to generate the control identifiers in the responses, signaling the intention of the response. Given the input intention signaled by the control identifiers, the user or agent is prompted to select a proper intention and generate responses accordingly.

\subsubsection{Simulation Details}
As described in Sec. \ref{Overview}, the simulation begins by generating user profiles, which initializes the user requests. Following this, based on input intent from the preceding round, the simulation selects a user or agent response intent and then uses the corresponding prompt for dialogue generation. This selection process is governed by the predetermined logic listed below. The \textbf{Generate Chat} intent refers to the expressions of greetings and gratitude that are triggered only at the beginning or end of a conversation and are skipped in the subsequent list.

The conversation can be initiated by either the user or the agent. The following describes the detailed mechanism that triggers the user's intent.

\begin{itemize}[leftmargin=*,itemsep=1pt,partopsep=0pt,topsep=0pt,parsep=0pt]
\item \textbf{Conversation Starts}: triggers user's \textbf{Inform Requirement} intent. Using randomization, the user simulator is instructed to choose a task of interest along with several related preferences and then generate a corresponding request.
\item \textbf{Inquire} from the agent: triggers user's \textbf{Inform Requirement} intent, corresponding to the scenario to answer the follow-up questions from the agent. 
\item \textbf{Report Search Result} from the agent: if the user's preference has not been fully expressed, the user's \textbf{Inform Requirement} intent will be triggered. If no candidate meets the search criteria, this will trigger the user's \textbf{Update Requirement} intent. Otherwise, the presence of a single candidate will initiate the user's \textbf{Ask for Action} intent, while the discovery of multiple candidates will prompt the user's \textbf{Ask for Recommendation} intent.
\item \textbf{Recommendation} from the agent: the user will be prompted to select from (i)  \textbf{Inquire Propertied} intent for more information or (ii) \textbf{Ask for Action} intent to proceed to make transactions.
\item \textbf{Report Action Result} from the agent: if all the tasks in the user profile have been completed, \textbf{General Chat} between the user and agent will be triggered, and then the conversation terminates; Otherwise, \textbf{Inform Requirement} intent is triggered for a new task.
\end{itemize}

\noindent Below is the intent-triggering mechanism for the agent simulation. 

\begin{itemize}[leftmargin=*,itemsep=1pt,partopsep=0pt,topsep=0pt,parsep=0pt]
\item \textbf{Inform Requirement} from the user: the agent is prompted to check if all the required slot values have been collected. If not, \textbf{Inquire} intent will be triggered to generate follow-up questions; Otherwise, the agent will search based on the user's requirement, and then generate a response based on \textbf{Report Search Result} intent.
\item \textbf{Inquire Properties} from the user: triggers agent's \textbf{Answer} intent.
\item \textbf{Ask for Recommendation} from the user: the agent is prompted to select the top candidate and then generate the response based on \textbf{Recommendation} intent.
\item \textbf{Ask for Action} from the user: the agent is prompted to make transactions and then generate a response based on \textbf{Report Action Result} intent.
\end{itemize}

\subsubsection{Slot Extraction}
It's important to note that the agent simulator must verify that all necessary information has been gathered before initiating a search. To manage this, a slot tracking module is employed to keep track of both the required and filled slots. With the \textbf{Inform Requirement} prompt, the user simulator can simultaneously provide dialogue utterances and the corresponding filled slot values. However, there is a possibility that the conversation generated by GPT-4 might not align with the outcomes of slot filling. This discrepancy can lead to repeated or even endless query loops from the agent. To address this issue, a slot extraction model, backed by GPT-4, is utilized to ensure that the generated conversation matches the slot-filling results. If inconsistencies are found, the conversation must be regenerated to maintain coherence.

\subsubsection{Generation Diversity}



To obtain a high-quality DST model, it is essential to have dialogue data that encompasses a wide range of diversity. To ensure the data generated possesses this diversity, we manually created ten rewriting templates, which were then expanded into hundreds of templates by GPT-4. These rewriting templates serve as a post-processing tool to enhance the diversity of the user and agent responses.

The details about the rewriting templates and rewritten outputs are shown in Appendix~\ref{Templates for Booking Responses}.


\subsection{Two-stage Fine-tuning Strategy}
\label{Two-stage Fine-tuning Strategy}
Taking into account the discrepancy in distribution between GPT-4 generated and real dialogues, directly merging generated and real data could cause the resulting model to deviate from the true distribution. To address this issue, we have designed a two-stage fine-tuning approach. Initially, we fine-tuned the LLaMA 2 model using the generated dialogue data. Following this, we continue to fine-tune the model with real data. The first step enables the model to learn fundamental task-oriented dialogue patterns. The second step ensures that the model effectively bridges the gap between generated and real dialogues, aligning closely with the true distribution.


\section{Experiments}
\subsection{Datasets and Metrics}
We conduct all the experiments on MultiWOZ 2.2\footnote{\url{https://github.com/budzianowski/multiwoz/tree/master/data/MultiWOZ_2.2}} \cite{woz2.2} and MultiWOZ 2.4\footnote{\url{https://github.com/smartyfh/MultiWOZ2.4}}~\cite{woz2.4}. MultiWOZ \citep{budzianowski-etal-2018-multiwoz} has been extensively utilized for evaluating the performance of DST, including 8,438, 1,000, and 1,000 samples for training, dev, and test sets with multi-turn dialogues, which are collected by a Wizard-of-Oz~(WOZ) setup and encompass a diverse array of domains. MultiWOZ 2.2 dataset refines the annotations in dev and test sets of MultiWOZ 2.1 \citep{woz2.1}. MultiWOZ 2.4~\cite{woz2.4} is the latest refined version correcting all incorrect labels in dev and test sets. Following \citet{wu-etal-2019-transferable}, we remove the domains of `hospital' and `police' from both MultiWOZ2.2 and MultiWOZ2.4 datasets because they only appear a few times in the training set and never occur in the dev and test set. By using the MultiWOZ schema, nearly 8000 new dialogues are generated. The detailed statistics of MultiWOZ 2.2 and MultiWOZ 2.4 datasets and the generated dialogue data are demonstrated in Table \ref{tbl:dataset}.

We adopt Joint Goal Accuracy (JGA) as the evaluation metric, which is the primary metric for DST. JGA is defined as the proportion of dialogue turns in which all the key-values are correctly predicted.

\begin{table}[t]
\centering
\resizebox{\linewidth}{!}{
 \setlength{\tabcolsep}{.4mm}{
\begin{tabular}{l|ccc}

\toprule
\textbf{Metric} $\downarrow$ \textbf{Dataset} $\rightarrow$  & \textbf{2.2} & \textbf{2.4} & \textbf{Generated} \\
\midrule
\rule{0pt}{4pt}No. of domains & 8 & 7 & 5\\
\rule{0pt}{8pt}No. of dialogues  & 8,438 & 8,438 & 7,556 \\
\rule{0pt}{8pt}Total no. of turns & 113,556 & 113,556 & 102,602\\
\rule{0pt}{8pt}Avg. turns per dialogue  & 13.46 & 13.46 & 13.57\\
\rule{0pt}{8pt}Avg. tokens per turn & 13.13 & 13.38 & 17.01 \\
\rule{0pt}{8pt}No. of slots & 61 & 37 & 17\\
\rule{0pt}{8pt}Have schema description & Yes & Yes & - \\
\rule{0pt}{8pt}Unseen domains in test set & No & No & - \\
\bottomrule
\end{tabular}}}
\caption{Statistics of MultiWOZ~(2.2 and 2.4) and the generated dataset used for training in our experiments. 
}
\label{tbl:dataset}
\end{table}

\begin{table}[t]
\centering
\resizebox{.98\linewidth}{!}{
\begin{tabular}{lcc}
\toprule
\textbf{Models}  & \textbf{MultiWOZ 2.2}  &  \textbf{MultiWOZ 2.4}\\
\hline
TRADE & {45.40}     &   {55.05} \\ 
UniLM &  {54.25}    &   {-} \\
DS-DST  & {51.70}     &   {-} \\
{TripPy}  & {53.50} & {64.75} \\
{AG-DST}  &  {57.26}    &   {-} \\
{SDP-DST} &{57.60}    &   {-} \\
D3ST$_\text{Base}$  & {56.10}     &   {72.10} \\
D3ST$_\text{Large}$ & {54.20}     &  {70.80} \\
D3ST$_\text{XXL}$ & {58.70}     &  {75.90} \\
{SPACE-3} & {57.50}     &   {-} \\
MSP-L &57.70  &-\\
RefPyDST  &{-}  &65.20\\
Diable  &{56.48}    &   {70.46} \\
DDSA  &{-}    &   {75.58} \\
SPLAT &56.60  &-\\
MoNET &{-}    &   {76.02} \\
SSNet & 62.10     &   {-} \\  
TOATOD$_\text{Small}$ &  {61.92}  &  {-}\\
TOATOD$_\text{Base}$  &  {63.79}    &  {-}\\
\hline
{\text{LUAS}$_\text{R}$}  &  {\textbf{65.42}}  &  {\textbf{77.20}} \\
{\text{LUAS}$_\text{R+G}$}  & {\textbf{66.25}}  &  {\textbf{78.20}} \\
\bottomrule
\end{tabular}
}
\caption{Joint Goal Accuracy for DST results on MultiWOZ 2.2 and MultiWOZ 2.4 dataset. `-' denotes that the results are not reported in the original paper.}
\label{table: Main_Results}
\end{table}

\begin{table*}[t]
\centering
\resizebox{.8\linewidth}{!}{
\renewcommand{\arraystretch}{1.1}
\begin{tabular}{lccccc}
\toprule
\textbf{Metric} $\downarrow$ \textbf{Replaced Domain} $\rightarrow$ & \textbf{Attraction} & \textbf{Hotel} & \textbf{Restaurant} & \textbf{Taxi}  & \textbf{Train} \\ 
\hline
Replaced Dialogues  & 2538  & 3235  & 3666  & 1397  & 2840  \\
Replaced Turns & 13348 & 30402 & 25768 & 6662  & 33364 \\
Avg. replaced turns per dialogue  & 5.26  & 9.40  & 7.03  & 4.77  & 11.75 \\ 
Avg. tokens per replaced turn & 15.57 & 15.54 & 15.33 & 18.28 & 16.44 \\
Avg. slots per replaced user turn & 1.38  & 2.75  & 2.54  & 1.37  & 2.90  \\
\bottomrule
\end{tabular}
}
\caption{Substituting details for 5 domains of MultiWOZ 2.2.}
\label{table:replaceDetail}
\end{table*}

\begin{table*}[t]
\centering
\renewcommand{\arraystretch}{1.2}
\resizebox{.9\linewidth}{!}{
 \setlength{\tabcolsep}{1.1mm}{
\begin{tabular}{cccccc} 
\toprule
\textbf{Replaced Domain} & \textbf{Impact} & \textbf{JGA} ({\textbf{$\Delta$ }}) & \textbf{Slot Precision} ({\textbf{$\Delta$ }}) & \textbf{Slot Recall} ({\textbf{$\Delta$ }}) & \textbf{Slot F1} ({\textbf{$\Delta$ }}) \\
\hline
Base &  0\% & 65.42 & 95.47\% & 93.25\% & 94.35\%\\
\cline{1-6}
Attraction & 28.1\% &64.99 ($-$0.43) & 95.46\% ($-$0.01\%) & 92.93\% ($-$0.32\%)&  94.17\% ($-$0.18\%)\\
Hotel & 42.1\% & 64.28 ($-$1.13) & 95.22\% ($-$0.25\%) & 92.83\% ($-$0.42\%)& 94.01\% ($-$0.34\%)\\
Restaurant & 41.2\% & 64.61 ($-$0.81) & 95.44\% ($-$0.03\%) & 93.30\% ($+$0.05\%)& 94.36\% ($+$0.01\%)\\
Taxi & 9.1\% & 65.22 ($-$0.20) & 95.62\% ($+$0.15\%) & 92.91\% ($-$0.34\%)& 94.25\% ($-$0.10\%)\\
Train & 38.4\% & 64.23 ($-$1.19) & 95.59\% ($+$0.12\%) & 92.67\% ($-$0.58\%)& 94.11\% ($-$0.24\%)\\
\cline{1-6}
Averaged & 31.20\% & 64.67 ($-$0.75) &  95.47\% ($-$0.00\%) & 92.93\% ($-$0.32\%) & 94.18\% ($-$0.17\%) \\
\bottomrule
\end{tabular}}
}
\caption{JGA for substituting real data with generated data on MultiWOZ 2.2 dataset.}
\label{table:replaceResults}
\end{table*}

\begin{table*}[t]
\centering
\renewcommand{\arraystretch}{1.1}
\resizebox{.98\linewidth}{!}{
 \setlength{\tabcolsep}{1.1mm}{
\begin{tabular}{ccccccc} 
\toprule
\multirow{2}{*}{\textbf{Dataset}} & \multirow{2}{*}{\textbf{\makecell[c]{Real Data \\ Size}}} & \multirow{2}{*}{\textbf{{JGA}$_\text{R}$}} & \multirow{2}{*}{\textbf{$\text{JGA}_\text{R+G}$ ({\textbf{$\Delta$ }})}} & \multicolumn{3}{c}{\textbf{Slot}} \\
\cline{5-7}
& & & & \textbf{\makecell[c]{$\text{Precision}_\text{R+G}$ ({\textbf{$\Delta$ }})}}  & \textbf{\makecell[c]{ $\text{Recall}_\text{R+G}$ ({\textbf{$\Delta$ }})}}  & \textbf{\makecell[c]{$\text{F1}_\text{R+G}$ ({\textbf{$\Delta$ }})}}\\
\hline
\multirow{4}{*}{\makecell[c]{MultiWOZ \\ 2.2}} 
& 1000 & 58.77 & 63.06 (\textbf{$+$4.29}) & 95.06\% (\textbf{$+$0.69\%}) & 92.39\% (\textbf{$+$1.46\%}) & 93.70\% (\textbf{$+$1.08\%}) \\
& 2000 & 62.66 & 64.43 (\textbf{$+$1.77}) & 95.33\% (\textbf{$+$0.27\%}) & 92.90\% (\textbf{$+$0.53\%}) & 94.10\% (\textbf{$+$0.41\%}) \\
& 4000 & 64.01 & 65.84 (\textbf{$+$1.83}) & 95.55\% (\textbf{$+$0.13\%}) & 93.21\% (\textbf{$+$0.30\%}) & 94.37\% (\textbf{$+$0.22\%}) \\
& All & 65.42 & 66.25 (\textbf{$+$0.83}) & 95.61\% (\textbf{$+$0.14\%}) & 93.55\% (\textbf{$+$0.30\%}) & 94.57\% (\textbf{$+$0.22\%}) \\
\hline  
\multirow{4}{*}{\makecell[c]{MultiWOZ \\ 2.4}} 
& 1000 & 64.60 & 69.69 (\textbf{$+$5.09}) & 97.15\% (\textbf{$+$1.09\%}) & 94.59\% (\textbf{$+$0.58\%}) & 95.85\% (\textbf{$+$0.83\%}) \\
& 2000 & 72.15 & 75.58 (\textbf{$+$3.43}) & 97.67\% (\textbf{$+$0.59\%}) & 95.90\% (\textbf{$+$0.46\%}) & 96.78\% (\textbf{$+$0.52\%}) \\
& 4000 & 75.81 & 77.29 (\textbf{$+$1.48}) & 98.08\% (\textbf{$+$0.27\%}) & 96.12\% (\textbf{$+$0.16\%}) & 97.09\% (\textbf{$+$0.21\%}) \\
& All & 77.20 & 78.20 (\textbf{$+$1.00}) & 97.88\% (\textbf{$+$0.07\%}) & 96.46\% (\textbf{$+$0.22\%}) & 97.16\% (\textbf{$+$0.14\%}) \\
\hline
\end{tabular}
}}
\caption{JGA and Slot Performance for fine-tuning with different sizes of real data from MultiWOZ 2.2 and 2.4.}
\label{table:sampleResults}
\end{table*}

\subsection{Implementation Details}
The GPT-4 version used for simulation is gpt-4-1106-preview. As for the fine-tuning stage, 8 Nvidia A100 (80G) GPUs are utilized for supervised full-parameter tuning with pytorch's FSDP framework \citep{fsdp}. The base model is 7B version\footnote{\url{https://huggingface.co/meta-llama/Llama-2-7b-hf}} of LLaMA 2. For each fine-tuning stage, the learning rate is set to 2e-5 with the cosine scheduler~\cite{loshchilov2016sgdr}, and the batch size is set to 8 on each GPU. We utilize Adam optimizer~\cite{kingma2014adam} with $\beta_1$ = 0.9, $\beta_2$ = 0.999, and the warm-up ratio is set to 3\%. Both fine-tuning stages last around two hours. For inference, vLLM\footnote{\url{https://github.com/vllm-project/vllm}}~\cite{kwon2023efficient} is used. 

\subsection{Baselines}
To assess the efficacy of the generated dialogue data,  fine-tune LLaMA 2 solely using real data, referring to it as LUAS$_\text{R}$, which serves as a strong baseline. We also conduct comparisons between our model and other strong baselines, including TRADE~\citep{wu-etal-2019-transferable}, UniLM~\cite{dong2019unified}, DS-DST~\cite{ds-dst}, {TripPy}~\cite{Heck2020}, {AG-DST}~\cite{ag-dst}, {SDP-DST}~\cite{lee-etal-2021-dialogue}, D3ST~\cite{d3st}, {SPACE-3}~\cite{he2022unified}, MSP-L~\cite{ijcai2022p607}, RefPyDST~\cite{king2023diverse}, Diable~\cite{lesci-etal-2023-diable}, DDSA~\cite{yang2023multi}, SPLAT~\cite{bebensee-lee-2023-span}, MoNET~\cite{zhang-etal-2023-monet}, SSNet~\cite{atawulla2023slot}, TOATOD~\cite{bang-etal-2023-task}.

\subsection{Results for DST prediction}
The whole results are shown in Table \ref{table: Main_Results}, it needs to be pointed out that our model is primarily compared with the \textit{generation-based models}, because \textit{classification-based models} can utilize external knowledge, leading to unfair comparisons.
{LUAS}$_\text{R}$ is only fine-tuned on the real data, and {LUAS}$_\text{R+G}$ is fine-tuned on both real and generated data. From these results, we have the following observations:

(1)~On both MultiWOZ 2.2 and 2.4 datasets, the performance of LLaMA 2 fine-tuned on real data ({LUAS}$_\text{R}$) surpasses previous DST baselines. This outcome underscores the exceptional effectiveness of LLaMA 2.


(2)~Furthermore, the incorporation of additional generated data yields significant performance improvements, with enhancements of 0.83\% on MultiWOZ 2.2 and 1\% on MultiWOZ 2.4. This improvement emphasizes the important role of generated data in boosting overall model performance. As is shown in the next section, the gain from the generated data can be even larger in case the real dialogue data is of a smaller size. For example, the enhancement can be as large as from 4.29\% to 5.09\% if only 1,000 dialogue real data exists. Considering the challenge in dialogue data collection,
this result highlights the pragmatic significance of integrating generated data for DST development across domains.


\subsection{Results of Substituting Real Data with Generated Data}
In order to further validate the quality and effectiveness of the generated dialogue data, we conduct a data replacement experiment for different domains on MultiWOZ 2.2. In these experiments, all dialogue data segments related to a specific domain will be removed, and the newly generated data will be inserted at the removed location. After replacement, the new training set will consist of 1 domain with the generated data and 4 others with real data. The replacement details are shown in Table \ref{table:replaceDetail}.

The model is also trained on LLaMA 2 7B, the results are shown in Table \ref{table:replaceResults}, and the `($\Delta$)' denotes the difference between the results of real data and real data with 1 domain replaced with generated data. Statistically, the generated data on average affects 31.2\% of the training data, the test JGA decrease is from -0.2 to -1.19 with an average of -0.75, and the slot precision is on par with before with the recall drops by -0.32\% on average. Compared to the reduction in training data size, the decreases in JGA and slot performance are relatively minor, suggesting that using generated data can effectively adapt DST models to a new domain with decent accuracy.

In practical applications, our method for automated dialogue generation offers a fast way to develop dialogue systems in new domains, resulting in considerable savings in both time and cost.

\subsection{Analysis}
\subsubsection{{The Effect of Adding Generated Data to Real Data of Various Sizes}}
\begin{figure*}[t]
  \centering

\resizebox{.98\linewidth}{!}{
\includegraphics[width=\linewidth]{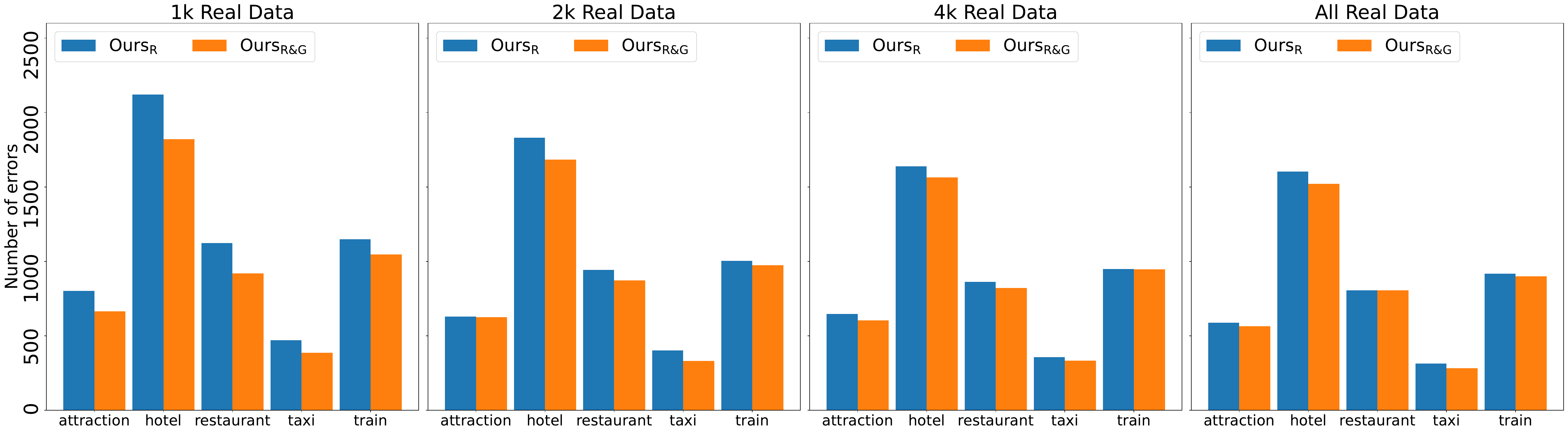}
}
\caption{The error distribution between $\text{LUAS}_\text{R}$ and $\text{LUAS}_\text{R+G}$ with different sizes of real data on MultiWOZ 2.2.}
\label{fig:error}
\end{figure*}

To better illustrate the impact of generated data, we conduct a series of experiments by combining generated data with various sizes of real data. 
The experiment results are demonstrated in Table \ref{table:sampleResults} and the sizes of real data used are set to be 1000, 2000, 4000, and all.
The $\text{JGA}_\text{R}$ represents the results obtained from training solely with real data, while $\text{JGA}_\text{R+G}$ and Slot $\text{Precision}_\text{R+G}$, $\text{Recall}_\text{R+G}$, and $\text{F1}_\text{R+G}$ represent the DST and slots accuracy results obtained from training with the same real data along with additional generated data. The symbols used in Table  \ref{table:replaceResults} are also used here.

The findings indicate that incorporating generated data into the training process significantly enhances model performance, surpassing that achieved with solely real data, particularly in scenarios where real training data is scarce.  
Under such situation, the performance of a model trained with generated data can be comparable to the model trained with twice amount of real data. For example, when only using 1,000 real data, the JGA of the two datasets will increase by 4.29\% and 5.09\% if the generated data is used, which is comparable to the performance of using 2,000 real data. Such findings hold considerable practical relevance, as they underscore the capacity of generated data to substantially mitigate the limitations posed by insufficient original data in real-world contexts.


\subsubsection{Error Distribution Analysis}
As illustrated in Figure \ref{fig:error}, to further highlight the superiority of our approach, we examine the error distribution of different sizes of real data on MultiWOZ 2.2 between $\text{LUAS}_\text{R}$ and $\text{LUAS}_\text{R+G}$. Using generated data leads to a reduction in errors across almost all domain categories compared to models fine-tuned solely on original data. This finding not only confirms the high quality of our generated data but also emphasizes the effectiveness of our approach in enhancing model performance in DST.
\section{Conclusion}
In this paper, we propose a novel approach utilizing GPT-4 to simulate conversations between users and agents and generate dialogues with dialogue state labels. We then conduct a two-stage fine-tuning of LLaMA 2 on both the generated and real data for DST prediction. Experimental results on two public DST benchmarks demonstrate that our model, augmented with generated data, outperforms the baseline trained solely on real data. Furthermore, detailed analysis confirms the adaptability of our approach, effectively meeting the dynamic requirements of transitioning to new domains in real-world scenarios. We believe that our method can be extended into a generalizable framework, offering benefits to a wide range of dialogue-related tasks.

\section*{Limitations}
With the high-quality dialogue data produced by our algorithm, we have significantly enhanced the DST model. We are also confident that the data generated will serve as valuable resources for other dialogue-related tasks, such as dialogue generation. We plan to explore this aspect in future research.
\section*{Ethics Statement}

We conduct experiments using two publicly available datasets  in addition to datasets created by GPT-4, with a specific focus on multi-domain task-oriented dialogue. Each dataset is subjected to thorough pre-processing for academic research purposes, which includes the removal of any personally identifiable information or offensive content. As a result, we are confident that our work presents no ethical risks.

\bibliography{custom}

\appendix
\clearpage

\section{Prompts for Simulation}
\label{Prompts for Simulation}

In this section, we illustrated a variety of typical prompts utilized by the user simulator and the agent simulator within the simulation. The symbol `$\backslash$n' of the prompts represents a line break.

Table \ref{table:prompt_for_user_inform_requirement} represents the prompt that the user informs requirement to the agent.
Table \ref{table:prompt_for_user_update_requirement} represents the prompt that the user updates the requirement to the agent.
Table \ref{table:prompt_for_user_to_ask_a_recommendation} represents the prompt that the user asks for a recommendation with the control identifier `\textbf{[RECOM]}' to request a recommendation from the agent. 
Table \ref{table:prompt_for_user_to_inquire_properties} represents the prompt that the user inquires about some properties (e.g. address and postcode) for the candidates.
Table \ref{table:prompt_for_user_to_active_a_booking_action} represents the prompt that the user asks for a booking action from the agent.
Table \ref{table:prompt_for_user_to_chat} represents the prompt that the user's general chat with the agent with the control identifier `[\textbf{EOF}]' to inform the agent that all the needs are satisfied.

Table \ref{table:prompt_for_slot_extracotr} represents the prompt that is used by the slot extractor.

Table \ref{table:prompt_for_agent_to_inquire_requirements} represents the prompt that the agent inquires the user for a specified requirement(e.g. restaurant-pricerange).
Table \ref{table:prompt_for_agent_to_answer_the_uers_inquiry_properties} represents the prompt that the agent responds with the properties that the user inquiries.
Table \ref{table:prompt_for_agent_to_report_search_result} represents the prompt that the agent reports search results to the user.
Table \ref{table:prompt_for_agent_to_report_action_result} represents the prompt that the agent reports the action result (e.g. the reservation information, etc.) and control identifier `\textbf{[BOOKED]}' to inform the successful of the reservation.
Table \ref{table:prompt_for_agent_to_complete_conversion} represents the prompt that the agent general chats with the user and outputs the control identifier `\textbf{[EOD]}' to end the whole simulation.

\begin{table*}[h]
\centering
\resizebox{.9\linewidth}{!}{
\begin{tabular}{p{\textwidth}}
\toprule
DST: [\textbf{\textit{history}}], [\textbf{\textit{requirements}}] $\rightarrow$ [\textbf{\textit{user\_utterance}}]   \\
\hline
\textbf{Prompt: }
This is your first time in Cambridge and want to find a restaurant.$\backslash$n
Now you are chatting with a local guide online.$\backslash$n 
And this is your preference:$\backslash$n
\{"restaurant's area": "north"\}$\backslash$n
and the conversation history (may be empty or not relevant to the current preference):$\backslash$n
{}[]$\backslash$n
Your responses should resemble an online chat as much as possible, and make them as brief as you can.$\backslash$n
How would you initiate the inquiry or respond to the guide online?$\backslash$n
Please do not provide any information that does not exist in your preference.$\backslash$n
Please randomly use synonyms or synonymous phrases to describe your intention, for example:$\backslash$n
- you can use `something to eat` or some food` instead of `restaurant`.Please provide all the information in your preferences to the guide, except the ones that have been informed in the history.$\backslash$n
Please remember not to provide any information that does not exist in your preference.$\backslash$n
If the local guide asks your preference, answer it directly and don't answer with other words.$\backslash$n 
Please don't repeat asking the same thing that is in the history.$\backslash$n
Please don't repeat your old preference which you have informed in the history when you respond to the guide.$\backslash$n
Please make sure the time in your response must be in the format `after, at or around \%H:\%M` in 24 hours.$\backslash$n
Pay attention to the diversity of responses, and try not to reuse sentence patterns that have been used in history.$\backslash$n
Only output the newest utterance, don't output the conversation history. \\
\textbf{Output: }
Do you know of any good eateries in the north of Cambridge?\\
\bottomrule
\end{tabular}}
\caption{Prompt for the user to inform requirement.}
\label{table:prompt_for_user_inform_requirement}
\end{table*}

\begin{table*}[h]
\centering
\resizebox{.9\linewidth}{!}{
\begin{tabular}{p{\textwidth}}
\toprule
DST: [\textbf{\textit{history}}], [\textbf{\textit{old requirements}}], [\textbf{\textit{new requirements}}] $\rightarrow$ [\textbf{\textit{user\_utterance}}]   \\
\hline
\textbf{Prompt: }
You are the first time in Cambridge and want to find a hotel.$\backslash$n
And now you are chatting with a local guide online.$\backslash$n
Here is your old preference:$\backslash$n
\{"hotel's area": "north", "hotel's stars": "4", "hotel's type": "hotel"\}$\backslash$n
and here is your new perference:$\backslash$n
\{"hotel's type": "guesthouse"\}$\backslash$n
and the conversation history:$\backslash$n
["Could you suggest any 4-star hotels in the northern part of Cambridge?", "from local guide: Are you looking for a boutique hotel or would a chain hotel suit your needs?", "A regular hotel would be fine, not looking for a boutique.", "from local guide: I apologize, but it seems we currently do not have any 4-star hotels available in the northern part of Cambridge matching your search criteria. Is there anything else I can assist you with or would you like to adjust your search conditions?"]$\backslash$n
Please output your response to inform the local guide of your preference change.$\backslash$n
Your responses should resemble an online chat as much as possible, and make them as brief as you can.$\backslash$n
Don't tell the guide you change your mind, please inform him like:$\backslash$n
- how about, would you like or do you have and ect.$\backslash$n
Only output the newest utterance, don't output the conversation history. \\
\textbf{Output: }
How about guesthouses? Do you have any 4-star options in the north of Cambridge for a group of 4, staying from Saturday for 4 nights? \\
\bottomrule
\end{tabular}}
\caption{Prompt for the user to update requirement.}
\label{table:prompt_for_user_update_requirement}
\end{table*}

\begin{table*}[h]
\centering
\renewcommand{\arraystretch}{1.1}
\resizebox{.9\linewidth}{!}{
\begin{tabular}{p{\textwidth}}
\toprule
DST: [\textbf{\textit{history}}], [\textbf{\textit{requirements}}] $\rightarrow$ [\textbf{\textit{user\_utterance}}] \\
\hline
\textbf{Prompt: }
This is your first time here and want to find a place to eat.$\backslash$n
Now you are chatting with a local guide online.  $\backslash$n
And this is your preference: $\backslash$n
$\backslash$n{"restaurant's area": "north", "restaurant's pricerange": "moderate", "restaurant's bookpeople": "8", "restaurant's booktime": "18:00"$\backslash$n} $\backslash$n
and the conversation history (may be empty or not relevant to the current preference): $\backslash$n
{}[ $\backslash$n
  "Do you know of any good eateries in the north of Cambridge?", $\backslash$n
  "from local guide: Are you looking for something more upscale or casual?", $\backslash$n
  "I'd like a recommendation for a place that's not too expensive.", $\backslash$n
  "from local guide: I found a couple of places that fit your preferences. Would you like me to recommend one?" $\backslash$n
] $\backslash$n
Your responses should resemble an online chat as much as possible, and make them as brief as you can.$\backslash$n
How would you initiate the inquiry or respond to the guide online? $\backslash$n
Please do not provide any information that does not exist in your preference.$\backslash$n
There are several choices that meet your preference.$\backslash$n
If the agent doesn't recommend you a selection,  $\backslash$n
please ask directly for a recommendation from the local agent, and output a special mark `[RECOM]` if you are looking for a recommendation.$\backslash$n
Only output the newest utterance, don't output the conversation history.\\
\textbf{Output: } Yes, please recommend one. [RECOM] \\
\bottomrule
\end{tabular}}
\caption{Prompt for the user to ask a recommendation with control identifier `\textbf{[RECOM]}'.}
\label{table:prompt_for_user_to_ask_a_recommendation}
\end{table*}

\begin{table*}[h]
\centering
\renewcommand{\arraystretch}{1.1}
\resizebox{.9\linewidth}{!}{
\begin{tabular}{p{\textwidth}}
\toprule
DST: [\textbf{\textit{history}}], [\textbf{\textit{requirements}}], [\textbf{\textit{properties}}]  $\rightarrow$ [\textbf{\textit{user\_utterance}}] \\
\hline
\textbf{Prompt: }
This is your first time in here and want to find a place to eat.$\backslash$n
Now you are chatting with a local guide online.$\backslash$n
And this is your preference:$\backslash$n
\{"restaurant's area": "north", "restaurant's pricerange": "moderate", "restaurant's bookpeople": "8", "restaurant's booktime": "18:00", "restaurant's bookday": "monday"\}$\backslash$n
and the conversation history (may be empty or not relevant to the current preference):$\backslash$n
{}[$\backslash$n
  "Do you know of any good eateries in the north of Cambridge?",$\backslash$n
  "from local guide: Are you looking for something more upscale or casual?",$\backslash$n
  "I'd like a recommendation for a place that's not too expensive.",$\backslash$n
  "from local guide: I found a couple of places that fit your preferences. Would you like me to recommend one?",$\backslash$n
  "I'd appreciate a recommendation.",$\backslash$n
  "from local guide: golden wok is an chinese food restaurant in the moderate price rang and the north part of town."$\backslash$n
]$\backslash$n
Your responses should resemble an online chat as much as possible, and make them as brief as you can.$\backslash$n
How would you initiate the inquiry or respond to the guide online?$\backslash$n
Please do not provide any information that is not exist in your preference.$\backslash$n
Here is some information that you want to get from the local guide:$\backslash$n
{}["address", "postcode"]$\backslash$n
Please read the history carefully and ask the information that is in your list but has not been mentioned in the history.$\backslash$n
Please ask a question for the information only, don't respond with other thing.$\backslash$n
Please try not to mention names in your questions as much as possible.$\backslash$n
Only output the newest utterance, don't output the conversation history.\\

\textbf{Output: } Could you provide the address and postcode for that place? \\
\bottomrule
\end{tabular}}

\caption{Prompt for the user to inquire the properties of the candidate.}
\label{table:prompt_for_user_to_inquire_properties}
\end{table*}

\begin{table*}[h]
\centering
\renewcommand{\arraystretch}{1.1}
\resizebox{.9\linewidth}{!}{
\begin{tabular}{p{\textwidth}}
\toprule
DST: [\textbf{\textit{history}}], [\textbf{\textit{requirements}}]  $\rightarrow$ [\textbf{\textit{user\_utterance}}] \\
\hline
\textbf{Prompt: }
Now you are chatting with a local guide online.$\backslash$n
And this is your preference:$\backslash$n
\{"hotel's area": "north", "hotel's stars": "4", "hotel's type": "guesthouse", "hotel's bookday": "saturday", "hotel's bookpeople": "4", "hotel's bookstay": "4"\}$\backslash$n
and the conversation history (may be empty or not relevant to the current preference):$\backslash$n
{}[$\backslash$n
  "Could you suggest any 4-star hotels in the northern part of Cambridge?",$\backslash$n
  "from local guide: Are you looking for a boutique hotel or would a chain hotel suit your needs?",$\backslash$n
  "A regular hotel would be fine, not looking for a boutique.",$\backslash$n
  "from local guide: I apologize, but it seems we currently do not have any 4-star hotels available in the northern part of Cambridge matching your search criteria. Is there anything else I can assist you with or would you like to adjust your search conditions?",$\backslash$n
  "How about guesthouses? Do you have any 4-star options in the north of Cambridge for a group of 4, staying from Saturday for 4 nights?",$\backslash$n
  "from local guide: We have several 4-star guesthouses available in the north of Cambridge that meet your criteria. Would you like me to recommend one?",$\backslash$n
  "Yes, please recommend one. ",$\backslash$n
  "from local guide: I currently have access to the worth house; how about we set up a reservation for you?",$\backslash$n
  "What's the postcode for the location?",$\backslash$n
  "from local guide: The Worth House is located at the postcode CB41DA. Would you like to proceed with the booking for 4 people from Saturday for a 4-night stay?"$\backslash$n
]$\backslash$n
Your responses should resemble an online chat as much as possible, and make them as brief as you can.$\backslash$n
How would you initiate the inquiry or respond to the guide online?$\backslash$n
Please do not provide any information that is not exist in your preference.$\backslash$n
Please ask for a booking from the local guide with your booking preference.$\backslash$n
Please don't use today or other relative days to describe the `bookday`.$\backslash$n
If no booking is needed, please end the conversion directly.$\backslash$n
If the guide asks you for the booking information, please avoid providing the booking information only.$\backslash$n
Please don't put other references that are non-relevant to your booking, like price range, area or others.$\backslash$n
Please try not repeat the booking information that you have already informed in the history.$\backslash$n
Only output the newest utterance, don't output the conversation history.\\
\textbf{Output: } Yes, please proceed with the booking at Worth House. \\
\bottomrule
\end{tabular}}
\caption{Prompt for the user to ask for a booking action from the agent.}
\label{table:prompt_for_user_to_active_a_booking_action}
\end{table*}

\begin{table*}[h]
\centering
\renewcommand{\arraystretch}{1.1}
\resizebox{.9\linewidth}{!}{
\begin{tabular}{p{\textwidth}}
\toprule
DST: [\textbf{\textit{history}}], [\textbf{\textit{requirements}}]  $\rightarrow$ [\textbf{\textit{user\_utterance}}] \\
\hline
\textbf{Prompt: }
This is your first time in cambridge and want to find a hotel.$\backslash$n
Now you are chatting with a local guide online. $\backslash$n
And this is your preference:$\backslash$n
\{"hotel's type": "guesthouse", "hotel's stars": "4", "hotel's internet": "yes", "hotel's area": "centre", "hotel's name": "alexander bed and breakfast", "hotel's parking": "yes", "hotel's bookstay": "1", "hotel's pricerange": "cheap", "hotel's bookday": "tuesday", "hotel's bookpeople": "6"\}$\backslash$n
and the conversation history (may be empty or not relevant to the current preference):$\backslash$n
{}[$\backslash$n
  "I'm eager to see 'the man on the moon' during my visit to Cambridge. Can you provide some help with this?",$\backslash$n
  "from local guide: Absolutely, I'd be happy to help with your visit to 'the man on the moon' in Cambridge. You can find this concert hall at 2 Norfolk Street, in the centre area.  ",$\backslash$n
  "Many thanks for the concert hall information. Could you point me to a good 4-star guesthouse close by?",$\backslash$n
  "from local guide: Would you prefer a guesthouse with free Wi-Fi access, and should it be in a specific area of Cambridge?",$\backslash$n
  "Yes, a guesthouse with internet access in the central area would be ideal.",$\backslash$n
  "from local guide: I've found a lovely 4-star guesthouse for you: the Alexander Bed and Breakfast, located centrally at 56 Saint Barnabas Road, with the convenience of free internet access. Would you like me to book a room for you?",$\backslash$n
  "That's just right. Can you book one night for me and ensure parking is covered?",$\backslash$n
  "from local guide: Fantastic choice! To proceed with the booking at Alexander Bed and Breakfast, could you please provide me with the date you'd like to stay and the number of people the booking is for?",$\backslash$n
  "I'd like to book a single room for the night of this coming Tuesday.",$\backslash$n
  "from local guide: Your single room at Alexander Bed and Breakfast for the night of this coming Tuesday has been successfully booked. Your reference code is: 3F7H8J9Z.  "$\backslash$n
]$\backslash$n
Your responses should resemble an online chat as much as possible, and make them as brief as you can.$\backslash$n
How would you initiate the inquiry or respond to the guide online?$\backslash$n
Please do not provide any information that is not exist in your preference.$\backslash$n
And now, all your preferences are met.$\backslash$n
Please always answer with **No** if the guide asks you whether more information is needed.$\backslash$n
Please thanks for the help from the local guide and output the mark `[EOF]`Only output the newest utterance, don't output the conversation history.\\
\textbf{Output: } Grateful for your guidance.[EOF] \\
\bottomrule
\end{tabular}}
\caption{Prompt for the user to chat with the agent with the control identifier `[\textbf{EOF}]'.}
\label{table:prompt_for_user_to_chat}
\end{table*}

\begin{table*}[h]
\centering
\resizebox{.9\linewidth}{!}{
\begin{tabular}{p{\textwidth}}
\toprule
DST: [\textbf{\textit{history}}], [\textbf{\textit{slots schema}}] $\rightarrow$ [\textbf{\textit{slots}}] \\
\hline
\textbf{Prompt: }You are a local agent, and now chatting with the user online for `restaurant.$\backslash$n
Here is the conversion history:$\backslash$n
{}["Do you know of any good eateries in the north of Cambridge?"]$\backslash$n
Here are the service schemas that you might use for all services:$\backslash$n
\{"restaurant": [\{"name": "restaurant-pricerange", "description": "price budget for the restaurant", "possible\_values": ["cheap", "expensive", "moderate"], "is\_categorical": true\}, \{"name": "restaurant-area", "description": "area or place of the restaurant", "possible\_values": ["centre", "east", "north", "south", "west"], "is\_categorical": true\}, \{"name": "restaurant-food", "description": "the cuisine of the restaurant you are looking for", "possible\_values": ["italian", "international", "indian", "chinese", "modern european", "european", "british", "gastropub", "mexican", "lebanese", "vietnamese", "spanish", "french", "japanese", "portuguese", "korean", "turkish", "asian oriental", "african", "mediterranean", "seafood", "thai", "north american"], "is\_categorical": true\}, \{"name": "restaurant-name", "description": "name of the restaurant", "possible\_values": [], "is\_categorical": false\}, \{"name": "restaurant-bookday", "description": "day of the restaurant booking", "possible\_values": ["monday", "tuesday", "wednesday", "thursday", "friday", "saturday", "sunday"], "is\_categorical": true\}, \{"name": "restaurant-bookpeople", "description": "how many people for the restaurant reservation", "possible\_values": ["1", "2", "3", "4", "5", "6", "7", "8"], "is\_categorical": true\}, \{"name": "restaurant-booktime", "description": "time of the restaurant booking", "possible\_values": [], "is\_categorical": false\}]\}$\backslash$n
Please read the history and the service schemas carefully:$\backslash$n
- first find the best service matched for the last utterance,$\backslash$n
- then find the slots of restaurants from the conversion history based on the schema of the restaurant.$\backslash$n
Your response should be in JSON format: \{"slots": \{"slot key": "slot value"\}, "service": ""\},$\backslash$n
The service you selected must be in the schema.$\backslash$n
The slots in your output must be in the schema of your predicted `service`,$\backslash$n
- the `slot key` must be mentioned in the schema$\backslash$n
- the `slot value` should be mentioned in the schema `possible\_values` if the slot value is categorical or you need to extract its value exactly from the conversion history.\\
\textbf{Output: } \{"slots": \{"restaurant-area": "north"\}, "service": "restaurant"\}\\
\bottomrule
\end{tabular}}

\caption{Prompt for the slot extractor of restaurant domain.}
\label{table:prompt_for_slot_extracotr}
\end{table*}

\begin{table*}[h]
\centering
\resizebox{.9\linewidth}{!}{
\begin{tabular}{p{\textwidth}}
\toprule
DST: [\textbf{\textit{history}}], [\textbf{\textit{inquire\_requirements}}] $\rightarrow$ [\textbf{\textit{agent\_utterance}}], [\textbf{\textit{inquire\_requirements}}]\\
\hline
\textbf{Prompt: }
You are a local agent for `restaurant`, and are chatting with the user online.$\backslash$n
You are going to rhetorically question some search criteria to make the user request more clearly.$\backslash$n
Here is the conversion history:$\backslash$n
{}["Do you know of any good eateries in the north of Cambridge?"]$\backslash$n
and the rhetorical slots that you will ask: $\backslash$n
{}["restaurant-pricerange"]$\backslash$n
Please read the history and rhetorical slots carefully.$\backslash$n
Then generate a brief rhetorical response to continue the conversion.$\backslash$n
- the response should resemble an online chat as much as possible, and make them as brief as possible.$\backslash$n
- please ask by the rhetorical slots directly, don't respond with other words, don't tell the user that you are narrowing down the option.$\backslash$n
 - please try asking all the rhetorical slots that are provided in the rhetorical slots at once.$\backslash$n
- for the service `train`, no return ticket is preferred from the user, and all the users will be adults as a group when booking tickets, but you need still to ask how many people instead.$\backslash$n
Pay attention to the diversity of responses, and try not to reuse sentence patterns that have been used in history.$\backslash$n
Please answer in a JSON format, \{"response": "", "inquire\_requirements": []\}\\
\textbf{Output: } \{"response": "Are you looking for something more upscale or casual?", "inquire\_requirements": ["restaurant-pricerange"]\} \\
\bottomrule
\end{tabular}}

\caption{Prompt for the agent to inquire the user for a specified requirement.}
\label{table:prompt_for_agent_to_inquire_requirements}
\end{table*}

\begin{table*}[t!]
\centering
\renewcommand{\arraystretch}{1.1}
\resizebox{.9\linewidth}{!}{
\begin{tabular}{p{\textwidth}}
\toprule
DST: [\textbf{\textit{history}}], [\textbf{\textit{search condition}}], [\textbf{\textit{search results}}] $\rightarrow$ [\textbf{\textit{agent\_utterance}}] \\
\hline
\textbf{Prompt: }
You are a local agent, and now are chatting with the user online for restaurant.$\backslash$n
Given the conversion history and search condition, please read the history and search condition carefully.$\backslash$n
Then generate a proper response to answer the user demands.$\backslash$n
Here is the conversion history:$\backslash$n
{}["Do you know of any good eateries in the north of Cambridge?", "from local guide: Are you looking for something more upscale or casual?", "I'd like a recommendation for a place that's not too expensive.", "from local guide: I found a couple of places that fit your preferences. Would you like me to recommend one?", "I'd appreciate a recommendation.", "from local guide: golden wok is an chinese food restaurant in the moderate price rang and the north part of town.", "May I have the address and postcode for that location, please?"]$\backslash$n
the search condition: \{"restaurant": \{"restaurant-area": "north", "restaurant-pricerange": "moderate", "restaurant-name": "golden wok"\}\}$\backslash$n
the search results: [\{"address": "191 Histon Road Chesterton", "area": "north", "food": "chinese", "name": "golden wok", "phone": "01223350688", "postcode": "cb43hl", "pricerange": "moderate"\}]$\backslash$n
Your response must resemble an online chat as much as possible, and make them as brief as possible.$\backslash$n
If you have not introduce the candidate to the user, please:$\backslash$n
- Inform the user with the name, and ask the user whether he needs a booking.$\backslash$n
Or else if you are responding to a booking request, please make sure you know the following information:$\backslash$n
 - the information must be known before booking a restaurant are book-day, book-time and book-people.$\backslash$n
- you can ask these three attributes all at once or step by step.$\backslash$n
When all the book information, which are book day, book hour and book people are provided by the user, please respond with a confirm:$\backslash$n
- please inform the user that the booking is successful.$\backslash$n
- please output the name in your response, and other information like bookday, booktime, bookpeople are not necessary to inform.$\backslash$n
- please add an 8 character' reference code with numbers and letters in your response.$\backslash$n
- please output a mark `[BOOKED]` at the end of the response.$\backslash$n
- if the user informed you he doesn't need a booking or reservation at this moment or booking later. Please reply with good politely and shortly, and also output the mark `[BOOKED]`.$\backslash$n
Please answer in a JSON format \{"response": ""\} \\
\textbf{Output: } Certainly! Golden Wok is located at 191 Histon Road Chesterton, with the postcode CB43HL. Would you like me to make a reservation for you? \\
\bottomrule
\end{tabular}}
\caption{Prompt for the agent to answer the user's inquiry properties.}
\label{table:prompt_for_agent_to_answer_the_uers_inquiry_properties}
\end{table*}

\begin{table*}[t!]
\centering
\renewcommand{\arraystretch}{1.1}
\resizebox{.9\linewidth}{!}{
\begin{tabular}{p{\textwidth}}
\toprule
DST: [\textbf{\textit{history}}], [\textbf{\textit{search condition}}], [\textbf{\textit{search results}}] $\rightarrow$ [\textbf{\textit{agent\_utterance}}] \\
\hline
\textbf{Prompt: }
You are a local agent, and now are chatting with the user online for restaurant.$\backslash$n
Given the conversion history and search condition, please read the history and search condition carefully.$\backslash$n
Then generate a proper response to answer the user demands.$\backslash$n
Here is the conversion history:$\backslash$n
{}["Do you know of any good eateries in the north of Cambridge?", "from local guide: Are you looking for something more upscale or casual?", "I'd like a recommendation for a place that's not too expensive."]$\backslash$n
the search condition: \{"restaurant": \{"restaurant-area": "north", "restaurant-pricerange": "moderate"\}\}$\backslash$n
the search results: [\{"address": "7 Milton Road Chesterton", "area": "north", "food": "indian", "name": "the nirala", "phone": "01223360966", "postcode": "cb41uy", "pricerange": "moderate"\}, \{"address": "191 Histon Road Chesterton", "area": "north", "food": "chinese", "name": "golden wok", "phone": "01223350688", "postcode": "cb43hl", "pricerange": "moderate"\}]$\backslash$n
Your response must resemble an online chat as much as possible, and make them as brief as possible.$\backslash$n
If you have not inform the user of the result, please first inform the user of the result:$\backslash$n
- the information should have the number of candidates, please don't use the exact number, use many, several, some or others instead.$\backslash$n
- and it is also necessary to ask if the user needs a recommendation.$\backslash$n
If you have already informed the user of the result, and the user what a recommendation, please do the follows:$\backslash$n
Please recommend one candidate with the restaurant name and detailed information from the search results directly, don't repeat the user need.$\backslash$n
The detailed information needed contains restaurant area, pricerange, food type.$\backslash$n
- please don't output the candidate's details.$\backslash$n
Please answer in a JSON format \{"response": ""\} \\
\textbf{Output: } \{"response": "I found a couple of places that fit your preferences. Would you like me to recommend one?"\}\\
\bottomrule
\end{tabular}}
\caption{Prompt for the agent to report search results.}
\label{table:prompt_for_agent_to_report_search_result}
\end{table*}

\begin{table*}[t!]
\centering
\renewcommand{\arraystretch}{1.1}
\resizebox{.9\linewidth}{!}{
\begin{tabular}{p{\textwidth}}
\toprule
DST: [\textbf{\textit{history}}], [\textbf{\textit{search condition}}], [\textbf{\textit{search results}}] $\rightarrow$ [\textbf{\textit{agent\_utterance}}] \\
\hline
\textbf{Prompt: }
You are a local agent, and now are chatting with the user online for restaurant. $\backslash$n
Given the conversion history and search condition, please read the history and search condition carefully. $\backslash$n
Then generate a proper response to answer the user demands. $\backslash$n
Here is the conversion history: $\backslash$n
{}["Do you know of any good eateries in the north of Cambridge?", "from local guide: Are you looking for something more upscale or casual?", "I'd like a recommendation for a place that's not too expensive.", "from local guide: I found a couple of places that fit your preferences. Would you like me to recommend one?", "I'd appreciate a recommendation.", "from local guide: golden wok is an chinese food restaurant in the moderate price rang and the north part of town.", "May I have the address and postcode for that location, please?", "from local guide: Certainly! Golden Wok is located at 191 Histon Road Chesterton, with the postcode CB43HL. Would you like me to make a reservation for you?", "Kindly reserve seating for eight at Golden Wok on Monday, at 18:00 hours."] $\backslash$n
the search condition: \{"restaurant": \{"restaurant-area": "north", "restaurant-pricerange": "moderate", "restaurant-name": "golden wok", "restaurant-bookday": "monday", "restaurant-bookpeople": "8", "restaurant-booktime": "18:00"\}\} $\backslash$n
the search results: [\{"address": "191 Histon Road Chesterton", "area": "north", "food": "chinese", "name": "golden wok", "phone": "01223350688", "postcode": "cb43hl", "pricerange": "moderate"\}] $\backslash$n
Your response must resemble an online chat as much as possible, and make them as brief as possible. $\backslash$n
If you have not introduce the candidate to the user, please: $\backslash$n
- Inform the user with the name, and ask the user whether he needs a booking. $\backslash$n
Or else if you are responding to a booking request, please make sure you know the following information: $\backslash$n
 - the information must be known before booking a restaurant are book-day, book-time and book-people. $\backslash$n
 - you can ask these three attributes all at once or step by step. $\backslash$n
When all the book information, which are book day, book hour and book people are provided by the user, please respond with a confirm: $\backslash$n
- please inform the user that the booking is successful. $\backslash$n
- please output the name in your response, and other information like bookday, booktime, bookpeople are not necessary to inform. $\backslash$n
- please add an 8 character' reference code with numbers and letters in your response. $\backslash$n
- please output a mark `[BOOKED]` at the end of the response. $\backslash$n
- if the user informed you he doesn't need a booking or reservation at this moment or booking later. Please reply with good politely and shortly, and also output the mark `[BOOKED]`. $\backslash$n
Please answer in a JSON format \{"response": ""\} \\
\textbf{Output: } \{"response": "Your reservation at Golden Wok for eight people on Monday at 18:00 hours has been successfully made. Your reference code is 3FA7B6Z9. [BOOKED]"\}\\
\bottomrule
\end{tabular}}
\caption{Prompt for the agent to report reservation information of booking action with control identifier `\textbf{[BOOKED]}'.}
\label{table:prompt_for_agent_to_report_action_result}
\end{table*}

\begin{table*}[t!]
\centering
\renewcommand{\arraystretch}{1.1}
\resizebox{.9\linewidth}{!}{
\begin{tabular}{p{\textwidth}}
\toprule
DST: [\textbf{\textit{history}}]$\rightarrow$ [\textbf{\textit{agent\_utterance}}] \\
\hline
\textbf{Prompt: }
You are a local agent for `restaurant`, and are chatting with the user online.$\backslash$n
Give your a conversion history and please read the history carefully.$\backslash$n
Here is the conversion history:$\backslash$n
{}["Do you know of any good eateries in the north of Cambridge?", "from local guide: Are you looking for something more upscale or casual?", "I'd like a recommendation for a place that's not too expensive.", "from local guide: I found a couple of places that fit your preferences. Would you like me to recommend one?", "I'd appreciate a recommendation.", "from local guide: golden wok is an chinese food restaurant in the moderate price rang and the north part of town.", "May I have the address and postcode for that location, please?", "from local guide: Certainly! Golden Wok is located at 191 Histon Road Chesterton, with the postcode CB43HL. Would you like me to make a reservation for you?", "Kindly reserve seating for eight at Golden Wok on Monday, at 18:00 hours.", "from local guide: Your reservation at Golden Wok for eight people on Monday at 18:00 hours has been successfully made. Your reference code is 3FA7B6Z9.  ", "I appreciate your assistance!"]$\backslash$n
Then generate a casual response to continue or end the conversion if is necessary.$\backslash$n
The casual response should be:- highly related to the conversion history, and briefly enough.$\backslash$n
- resemble an online chat as much as possible, and make them as brief as possible.$\backslash$n
- make the reply simple when you respond to the users' thanks.$\backslash$n
- all the words in your response, should be limited to 15 words.$\backslash$n
- please also add a mark `[EOD]` at the end of your response.$\backslash$n
Please answer in a JSON format, \{"response": ""\} \\
\textbf{Output: }\{"response": "You're welcome! Enjoy your meal at Golden Wok. If you need anything else, just ask. [EOD]"\}\\
\bottomrule
\end{tabular}}

\caption{Prompt for the agent to conclude the dialogue with control identifier `\textbf{[EOD]}'.}
\label{table:prompt_for_agent_to_complete_conversion}

\end{table*}

\section{Templates for Booking Responses}
\label{Templates for Booking Responses}

We first crafted template responses like \textit{There are a lot of \{type\} attractions available. Would you like information about one of those? Perhaps, a \{type\} like \{name\}?} and then expand them to hundreds with the GPT-4's rewriting ability as shown in Table \ref{table:prompt for template rewriting}. In our simulation, the templated response will randomly substitute the recommendation responses of the agent to enhance the variety of interactions.

\begin{table*}[h]
\centering
\resizebox{.9\linewidth}{!}{
\begin{tabular}{p{\textwidth}}
\toprule
DST: [\textbf{\textit{template}}]$\rightarrow$ [\textbf{\textit{rewrited templates}}] \\
\hline
\textbf{Prompt: }$\backslash$n
Please rewrite the following template into 10 different ways,$\backslash$n
template:$\backslash$n
There are a lot of \{type\} attractions available. Would you like information about one of those? Perhaps, a \{type\} like \{name\}?$\backslash$n
Your output should be in JSON format like ["template0", "template1", ...]$\backslash$n
Your output:$\backslash$n  \\
\textbf{Output: }$\backslash$n
{}[$\backslash$n
  "Many \{type\} attractions are at your disposal. Interested in learning more about any? Maybe one such as \{name\}?",$\backslash$n
  "A wide array of \{type\} attractions are on offer. Care to get details on any specific one, possibly the \{name\}?",$\backslash$n
  "We have numerous \{type\} attractions to choose from. Would you care for details about a particular one, for instance, \{name\}?",$\backslash$n
  "Our selection of \{type\} attractions is quite extensive. Can I interest you in information about a specific one, maybe \{name\}?",$\backslash$n
  "You'll find plenty of \{type\} attractions here. Shall I provide information on one, perhaps \{name\}?",$\backslash$n
  "There's no shortage of \{type\} attractions to explore. Would you like to hear about one, such as \{name\}?",$\backslash$n
  "We boast a variety of \{type\} attractions. Would any pique your interest, maybe something like \{name\}?",$\backslash$n
  "The range of \{type\} attractions we have is impressive. Do you wish to know more about any, like \{name\}?",$\backslash$n
  "If you're interested in \{type\} attractions, we have many options. How about more information on one like \{name\}?",$\backslash$n
  "Looking for \{type\} attractions? We've got plenty. Want to learn about a specific one, say \{name\}?"$\backslash$n
]\\
\bottomrule
\end{tabular}}
\caption{Prompt for template rewriting.}
\label{table:prompt for template rewriting}
\end{table*}

\end{document}